\documentclass[aps,twocolumn,showpacs,amsmath,amssymb,prl
]{revtex4-1}

\usepackage{tikz}

\setcitestyle{numbers,square}
\usepackage{lmodern}		
\usepackage{cmap}			

\usepackage[utf8]{inputenc}
\usepackage{calc}
\usepackage{xifthen}
\usepackage[]{algorithm2e}

\usepackage{xr}
\usepackage{amssymb}
\usepackage{amsmath}
\usepackage{multirow}
\usepackage{todonotes}
\usepackage{graphics,graphicx}
\usepackage{xspace}
\usepackage{dsfont}
\usepackage{bm}
\usepackage{hyperref}
\usepackage{color}

\hypersetup{
	colorlinks,%
	citecolor=blue,%
	filecolor=blue,%
	linkcolor=blue,%
	urlcolor=blue
}

\newcommand{\rmig}{r_{mig}}

\newcommand{\mean}[2]{%
	\ifthenelse{\equal{#2}{}}{\left\langle #1 \right\rangle}{\left\langle #1 \right\rangle_{#2}}%
}

\newcommand{\spin}[1][]{%
	\ifthenelse{\equal{#1}{}}{s_{i\alpha}^{(ln)}}{s_{i#1}^{(ln)}}%
}
\newcommand{\spinc}[1][]{%
	\ifthenelse{\equal{#1}{}}{s_{i\alpha}}{s_{i#1}}%
}

\newcommand{\revfig}[1]{Fig.\ref{#1}}

\newcommand{\K}[2]{%
	\ifthenelse{\equal{#2}{}}{K_{#1}^{(ln)}}{K_{#1}^{(ln)(#2)}}%
}

\newcommand{\reveq}[1]{Eq.~(\ref{#1})}

\begin{document}

\title{A theory of multipopulation genetic algorithm with an application to the Ising model}

\author{Bruno Messias}
\email{messias@ifsc.usp.br}
\affiliation{Instituto de Física de São Carlos, Universidade de São Paulo,  São Carlos, SP 13566-590, Brazil}

\author{Bruno W. D. Morais}
\affiliation{Faculdade de Computação, Universidade Federal de Uberlândia,  Uberlândia, MG 38400-902, Brazil}
\date{\today}
\begin{abstract}
Genetic algorithm (GA) is a stochastic metaheuristic process consisting on the evolution of a population of candidate solutions for a given optimization problem. By extension, multipopulation genetic algorithm (MPGA) aims for efficiency by evolving many populations, or ``islands'', in parallel and performing migrations between them periodically. The connectivity between islands constrains the directions of migration and characterizes MPGA as a dynamic process over a  network. As such, predicting the evolution of the quality of the solutions is a difficult challenge, implying in the waste of computer resources and energy when the parameters are inadequate. 
By using models derived from statistical mechanics, this work aims to estimate equations for the study of dynamics in relation to the connectivity in MPGA.
To illustrate the importance of understanding MPGA, we show its application as an efficient alternative to the thermalization phase of Metropolis–Hastings algorithm applied to the Ising model.

\end{abstract}
\maketitle
\section{Introduction}
Genetic algorithm (GA) is a stochastic population-based technique used in search and optimization problems, with applications in fields like Computer Science, Engineering, Biology, and Physics\cite{Curtis2018,PhysRevB.97.100102, astronomiaGenetica2015, 2018structuresPrediction}. Aiming to achieve more time-efficiency on modern computers, Multipopulation Genetic Algorithm (MPGA)\cite{morady2016multi, lee1988multiprocessor, Bauke2003} is an approach for parallel and distributed modeling of GA.

MPGA can be described as a network of GA instances (islands) that evolve solutions semi-independently. Thus, MPGA can be understood as phenomenon of dynamics over a network. Besides time-efficiency, this modeling of islands and its resulting local interactions have an impact on the algorithm's search efficiency, which distinguishes MPGA as a different technique from GA \cite{Alba:1999:SPD:315491.315495}.

\begin{figure}[!htb]
    \centering
    \includegraphics[width=\columnwidth,keepaspectratio]{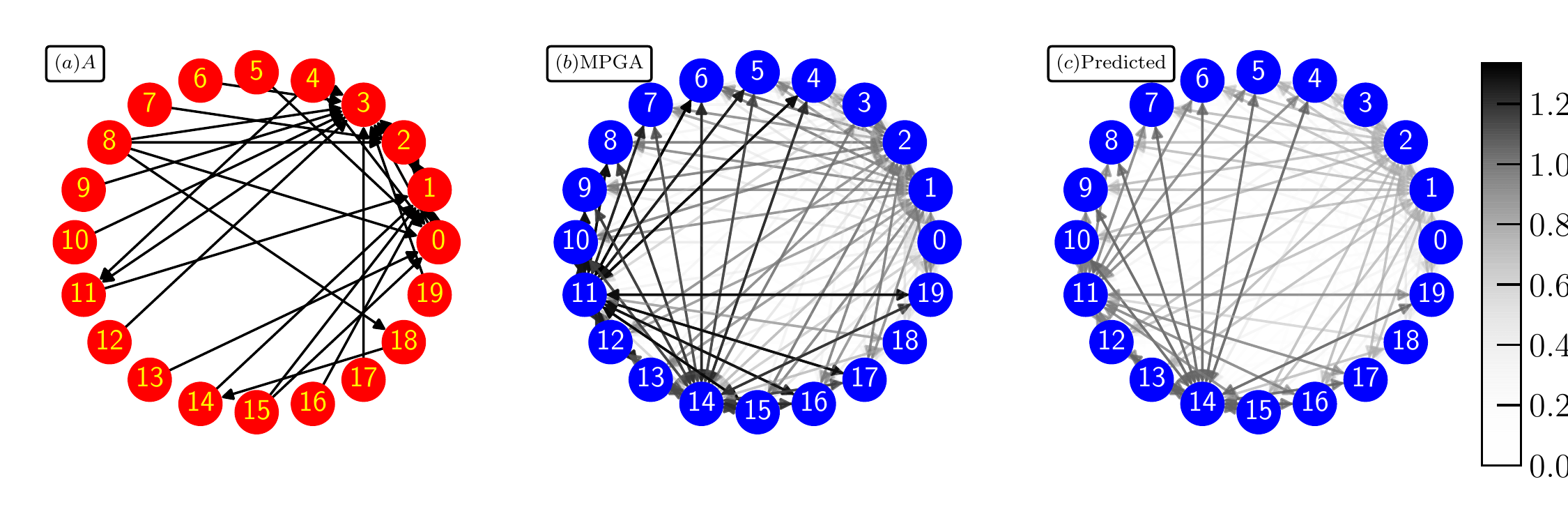}
    \caption{ 20-island MPGA for energy minimization of a $20$-spin ideal unidimensional paramagnet for temperature $\beta=-0.005$, with Boltzmann selection and without the crossover and mutation operators. Island connection, $\mathbf{A}$, is shown in $(a)$. Remaining parameters are $\Delta t_m =20$, $\rmig=0.2$, $N_P=100$. $(b)$ and $(c)$ show, respectively, the empirical and theoretical mappings of MPGA at generation $194$ to a weighted directed graph where weights are given by the Kullback-Leibler divergence between the islands' populations. }
    \label{figEvoTopology}
\end{figure}

In this work we provide tools to analyze the impact of the network connectivity on both the solutions and the behavior of the islands that constitute the GA. These tools are an extension of the cumulant dynamics formalism developed by Shapiro \cite{1994staticalga}, which provides useful insights about the behaviour of the distribution of individuals in MPGA, while other approaches are better suited for the analysis of run time bounds  \cite{Cantupaz2000, corus2017level, davis1993markov}.
To evaluate our methods, we developed a MPGA code for energy minimization of a unidimensional paramagnet.

The analysis of its dynamics enables a more effective development of MPGA regarding the usage of computational resources, and illustrates the rich phenomena that occur in it.

The improvement of MPGA becomes interesting in Physics when one realizes the optimization problems that arise in many of its subfields. As an application to Physics, we propose MPGA as an alternative approach to the thermalization phase of the Metropolis–Hastings algorithm (MH)\cite{metropolis1953equation, landau2014guide} applied to the Ising model. 
While practical, MH requires high usage of computer resources for problems with large configuration spaces, due to MH being a local search heuristic. In this context, previous works propose the improvement of the algorithm's efficiency\cite{acceleratedMonteCarlo, thomas2018}. While GA was proposed in the literature as an alternative to MH \cite{anderson1991two, maksymowicz1994genetic}, this is the first mention of MPGA/MH as an extension of it. Recently, the GA/MH approach was mentioned \cite{thomas2018}, where a GPU architecture was applied, and the study of different selection methods was suggested.

\section{Methods} Each MPGA island starts with a population of random candidate solutions (individuals), which is evolved iteratively over $N_g$ generations by creating new individuals and discarding ones of low quality (fitness). Individuals are created in a procedure called crossover, which combines two individuals (parents) to generate a new one. The selection of individuals for reproduction depends on their fitness.

Islands can be implemented as processes of the operational system. They communicate by sending individuals to each other (migration). An usual approach is to perform migration periodically in a regular interval of generations, ($\Delta t_{migr}$). The direction of migration (sender island to destination island) is given by a parameter of connectivity relation.

\begin{figure}
    \centering
    \includegraphics[width=\columnwidth,keepaspectratio]{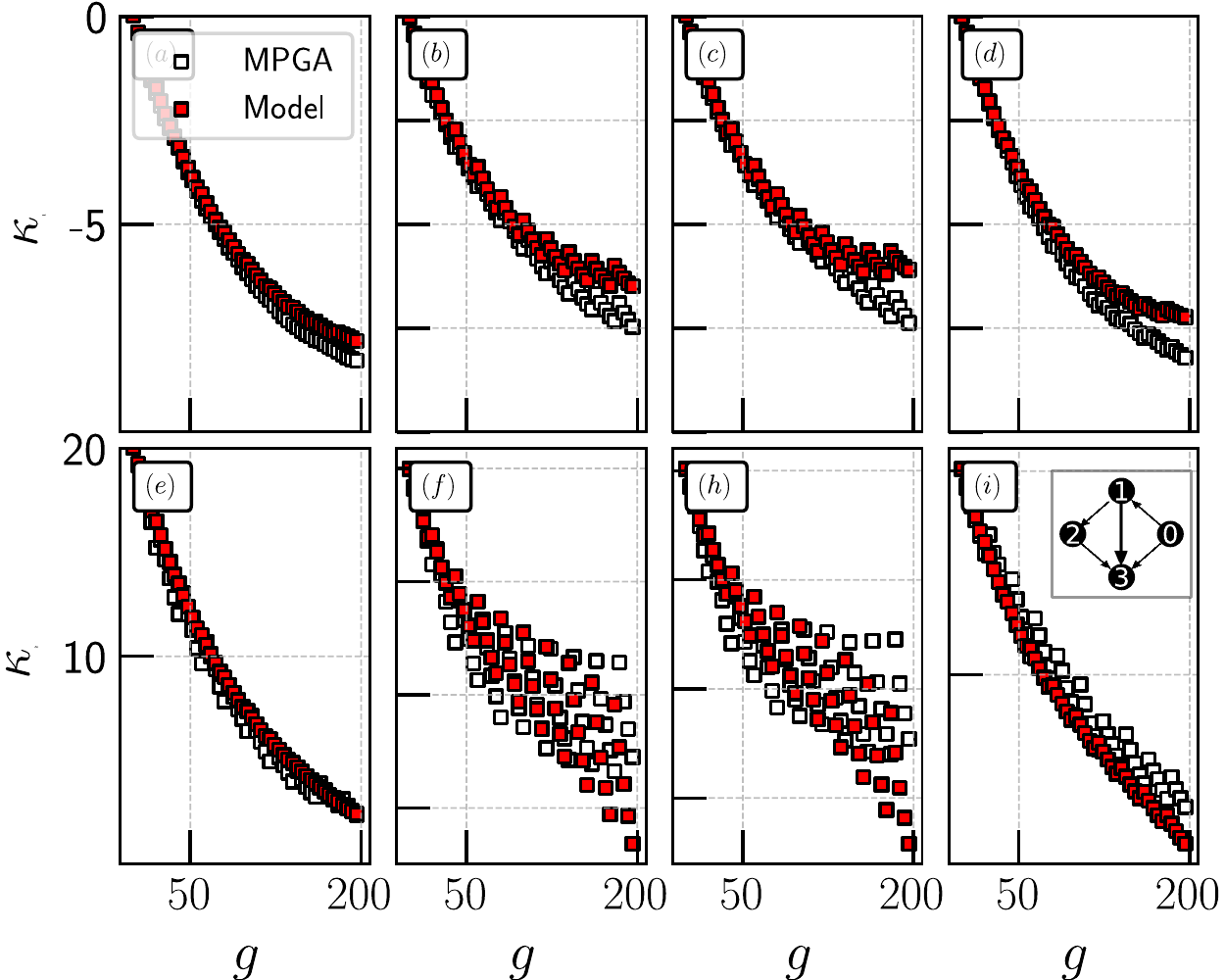}
    \caption{ First and second cumulants as a function of generation $g$ in the 4-island MPGA applied to energy minimization of a paramagnet of $20$ for temperature ($\beta=-0.005$), with Boltzmann selection and without crossover and mutation operators. Remaining parameters are $\Delta t_m = 20$, $\rmig = 0.2$, $N_P=100$. Red squares represent theoretical prediction considering the first three cumulants. White squares represent empirical results for $1000$ different executions of MPGA. Island connections are shown on inset (i). }
    \label{figModel_vs_Exp}
\end{figure}
The MPGA's island connectivity, exemplified in  \revfig{figEvoTopology} (a), is defined by an adjacency matrix $\mathbf A$ where $A_{ij}=1$ (or $0$) indicates that migrating individuals have non-zero (or zero) probability of moving from island $i$ to island $j$. Remaining parameters are: population size for each island ($N_P$), number of generations ($N_g$), migration period
($\Delta t_{mig}$), crossover rate ($r_{cross}$) and mutation rate ($r_{mut}$). 

To each population's individual is associated its fitness ($f$), which represents its quality according to a chosen criterion, e.g. minimization of a function. An individual's fitness is proportional to the probability of propagating its attributes along generations.

On a given island's population, values of $f$ can be used to define a probability function of $f$ by assuming that individuals are organized approximately in a gaussian distribution. Therefore, we can approximate the distribution of individuals with respect to $f$ with the Gram-Charlier expansion\cite{kendall1977advanced}. This expansion is obtained from the cumulant values $\kappa_i^{(ln)}$ for each island $l$ and each each generation $n$, resulting in the following probability function:

\begin{align}
p_{ln}(f) = \left(1+\Psi_{ln}\left(\scriptstyle{\frac{f - \kappa_1^{(ln)}}{\sqrt{\kappa_2^{(ln)}}}
}\right)\right) \frac{\exp\left[-\frac{( f - \kappa_1^{(ln)} )^2}{2\kappa_2^{(ln)}}\right]}{\sqrt{2\pi\kappa_2^{(ln)}}},
\end{align}
where
\begin{align}
\Psi_{ln}(x) = \sum\limits_{i=3}^\infty a_i^{(ln)} H_i(x) =  \sum\limits_{i=3}^\infty \frac{\kappa_i^{(ln)}}{i!\left(\kappa_2^{(ln)}\right)^{i/2}} H_i(x),
\end{align}
and $H_i(x)$ is the $i$-th probabilistic Hermite polynomial.

In a generation where migration occurs, each island's population is dependent on the others. In this case, there is a different set of cumulants $\{\tilde \kappa_i^{(ln)}\}$. To determine this set, we start defining that $r_{mig} N_P$ individuals migrate from each population. Hereafter, $\tilde A_{jl}=A_{jl}/\sum\limits_l A_{ji}$ is the normalized connection between islands $j$ and $l$, $n_m = 
r_{mig}\sum\limits_{j}\tilde A_{jl}$ is the rate of individuals that migrate to island $l$, $n_0 = 1-r_{mig}$ is the rate of individuals that stay on $l$. Let $n_r=r_{mig} - n_m$ be the rate of individuals to be generated to keep the population size equal to $N_P$. The two first cumulants for generation $n$ and island $l$ is given by 

\begin{align}
\tilde \kappa_1^{(ln)} 
&=
\frac{
    n_0\kappa_1^{(ln)}
    +r_{mig}\sum\limits_{j}\tilde A_{ji}\kappa_{1}^{(jn)}+n_r\Theta(n_r)\bar \kappa_{1}
}
{
    n_0
    +n_m
    +n_r\Theta(n_r),
} 
\nonumber \\
\tilde \kappa_{2}^{(ln)}
&= 
\frac{
    n_0\kappa_{2}^{(ln)}
    +r_{mig}\sum\limits_{j}\tilde A_{ji}\kappa_{2}^{(jn)}+n_r\Theta(n_r)\bar \kappa_{2}
}
{
    n_0
    +n_m
    +n_r\Theta(n_r),
} \nonumber \\
&+  \frac{
    n_0(\kappa_{1}^{(ln)})^2
    +r_{mig}\sum\limits_{j}\tilde A_{ji}(\kappa_{1}^{(jn)})^2+n_r\Theta(n_r)\bar \kappa_1^2
}
{
    n_0
    +n_m
    +n_r\Theta(n_r),
} 
\nonumber \\
&-(\bar\kappa_{1}^{(ln)})^2,
\label{eqCumuRede}
\end{align}
where $\Theta(n_r)$ is the step function and $\bar \kappa_{1}$ is the first cumulant extracted from a probability function, which is used to keep the population size invariant, if $\sum_j\tilde A_{jl}$ is small, $l$'s population size can get smaller than the $N_P$ after migration. To fill each island, new individuals are  generated randomly, which can have an effect on the local optimality of solutions.

With these probability functions defined for each island, we can analyze their evolution. Shapiro et al. \cite{1994staticalga} demonstrate how to determine the cumulant dynamics  using the formalism of random energy model \cite{derrida1981random}. In MPGA, the same model applies for migration, since selection is also applied to choose migrating individuals, with the addition of obtaining the first probability function by using the first cumulant, as described by equation \reveq{eqCumuRede}, and constructing the next cumulants using the first. In the Derrida-Shapiro model, cumulants' dynamic are determined by

\begin{align}
\kappa_m^{(l n+1)}= -\lim\limits_{\gamma\to 0}\frac{\partial^m}{\partial\gamma^m}
\int\limits_0^\infty\mathrm d  t\frac{(\int\limits_{-\infty }^\infty \mathrm d  f p_{ln}(f)\exp(-t\omega(f)e^{\gamma f})^{N_P}}{t},
\label{eqCumuShapiro}
\end{align}
where $\omega(f)$ is the function that defines the probability of selecting an individual with fitness $f$.

In MPGA, it is also interesting to analyze how the islands' populations differ from each other over the generations, and how their connections, given by the matrix $\mathbf A$, influence this dynamic. To model this, we present a mapping of MPGA to a weighted directed graph where nodes represent islands, edges represent their connections, and weights are given by the Kullback-Leibler divergence\cite{elementsInformationTheory}, $\mathbb{KL}(p_{ln}|| p_{qn})$.

For a weak enough selection, weights can be obtained with enough precision from the first two cumulants. Therefore, the Kullback-Leibler divergence $\mathbb{KL}(p_{ln}|| p_{qn})$ can be obtained from the gaussian distribution $\mathcal N(\kappa_1^{(ln)},\kappa_2^{(ln)})$ and the distribution given by 
$\mathcal N(\kappa_1^{(qn)}, \kappa_2^{(qn)})$.

In general, it is required to make corrections involving higher order cumulants. Assuming $\ln(1+\Psi_{ln}(x)) \approx \Psi_{ln}(x) -\frac{\Psi_{ln}(x)^2}{2}$, the correction term in relation to divergence between two gaussian distributions is given by

\begin{align}
\mathbb{\tilde{KL}}(p_{ln}|| p_{qn})
= 
\frac{(\kappa_4^{(ln)})^2}{48(\kappa_2^{(ln)})^4}+\frac{(\kappa_{3}^{(ln)})^2}{12 (\kappa_2^{(ln)})^3}
\nonumber \\
-\sum\limits_{j=3}^4a_j^{(qn)} \mu_{q_2}^{j/2}H_j(m_q)
\nonumber \\
- \frac{
    4 \sqrt{\kappa_2^{(ln)}} \kappa_3^{(ln)} \left(\sqrt{\kappa_2^{(qn)}} \kappa_3^{(qn)} \sqrt{\tilde q_2}-\kappa_4^{(qn)} \tilde q_1\right)+\kappa_4^{(ln)} \kappa_4^{(qn)}
}{
    24 (\kappa_2^{(ln)} \kappa_2^{(qn)} \tilde q_2)^2
}
\nonumber  \\
+\frac{1}{2}\sum\limits_{i,j=3}^4
\sum\limits_{k=0}^{\min (i, j)}a_{i}^{(ln)}a_{j}^{(qn)}
k!\mu_{q_2}^{\frac{i+j-2k}{2}} 
H_{i+j-2k}(m_q)\binom{j}{k}\binom{i}{k},
\label{eqKl2}
\end{align}
where

$\tilde{q_1} =  
\frac{\kappa_1^{(qn)}-\kappa_1^{(ln)}}{\sqrt{\kappa_2^{(ln)}}},
\ \ \tilde{q_2} = \frac{\kappa_2^{(qn)}}{\kappa_2^{(ln)}},
\ \ \mu_{q_2} = 1-\frac{1}{\tilde q_2},
$ e $m_{q} = -\frac{\tilde q_1}{\sqrt{\tilde q_2 -1}}$. 

\reveq{eqKl2} enables the mapping to a weighted directed graph that displays the dissimilarity between the islands' populations, although it can reach the limitations of the Kullback-Leibler divergence and Gram-Charlier expansion.

Having defined the theoretical framework, experiments with MPGA were developed by making use of the OpenMPI\cite{open_mpi} and MPI4Py\cite{dalcin2005mpi} libraries, which enabled the modeling of islands as communicating computer processes.

\section{Results}
 To validate \reveq{eqCumuRede}, \reveq{eqCumuShapiro} and the proposed mapping, we approach the problem of energy minimization of a system described by a paramagnet, with the absence of the crossover and mutation operators, since their effect were already discussed by Shapiro\cite{1994staticalga}. \revfig{figModel_vs_Exp} compares empirical results obtained by $1000$ MPGA experiments with theoretical estimates. This MPGA is composed by $4$ islands, where the probability of individual $\alpha$ from the $l$-th island being selected  is given by $e^{-\beta f_\alpha^{(ln)}}/\sum\limits_ie^{-\beta f_i^{(ln)}}$, where $\beta = 0.005$, which allow expand the \reveq{eqCumuShapiro}. Results demonstrate that an extension of Shapiro's proposed theoretical model is capable of covering the migration phenomenology. Peaks that arise during migration events are caused by random generation of individuals, who usually have poor fitness and don't propagate because of their low probability of selection. As can be noted, migration has strong effects on the second and higher cumulants, and this can be beneficial to MPGA by ensuring diversity as the islands evolve.

\begin{figure}[!htb]
    \centering
    \includegraphics[width=\columnwidth,keepaspectratio]{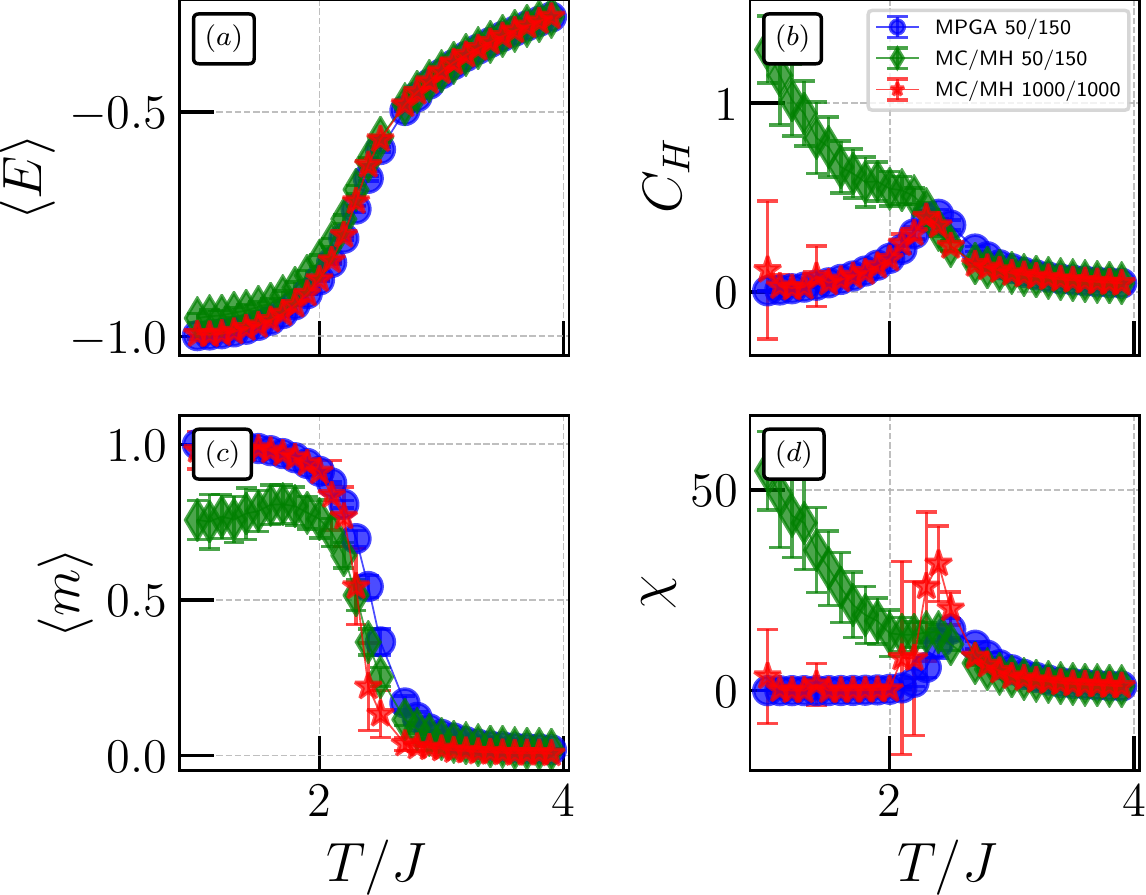}
    \caption{Mean energy, $\langle E\rangle$,  specific heat, $C_H$, magnetization, $\langle m \rangle$ and susceptibility $\chi$ as functions of temperature $T$ for Ising model with $20\times20$ spins at zero magnetic field and $J=1$. Each panel shows results respectively for: $6$-island MPGA ($N_P=20$ and $6$ experiments) 
        with $50$ generations and $150$ MH steps to calculate the mean values; MH with $50$ thermalization steps and $150$ calculation steps ($720$ experiments); MH with $1000$ thermalization steps and $1000$ calculation steps  ($12$ experiments). }
    \label{figMPGAvsMC}
\end{figure}
\begin{figure}[!htb]
    \centering
    \includegraphics[width=\columnwidth,keepaspectratio]{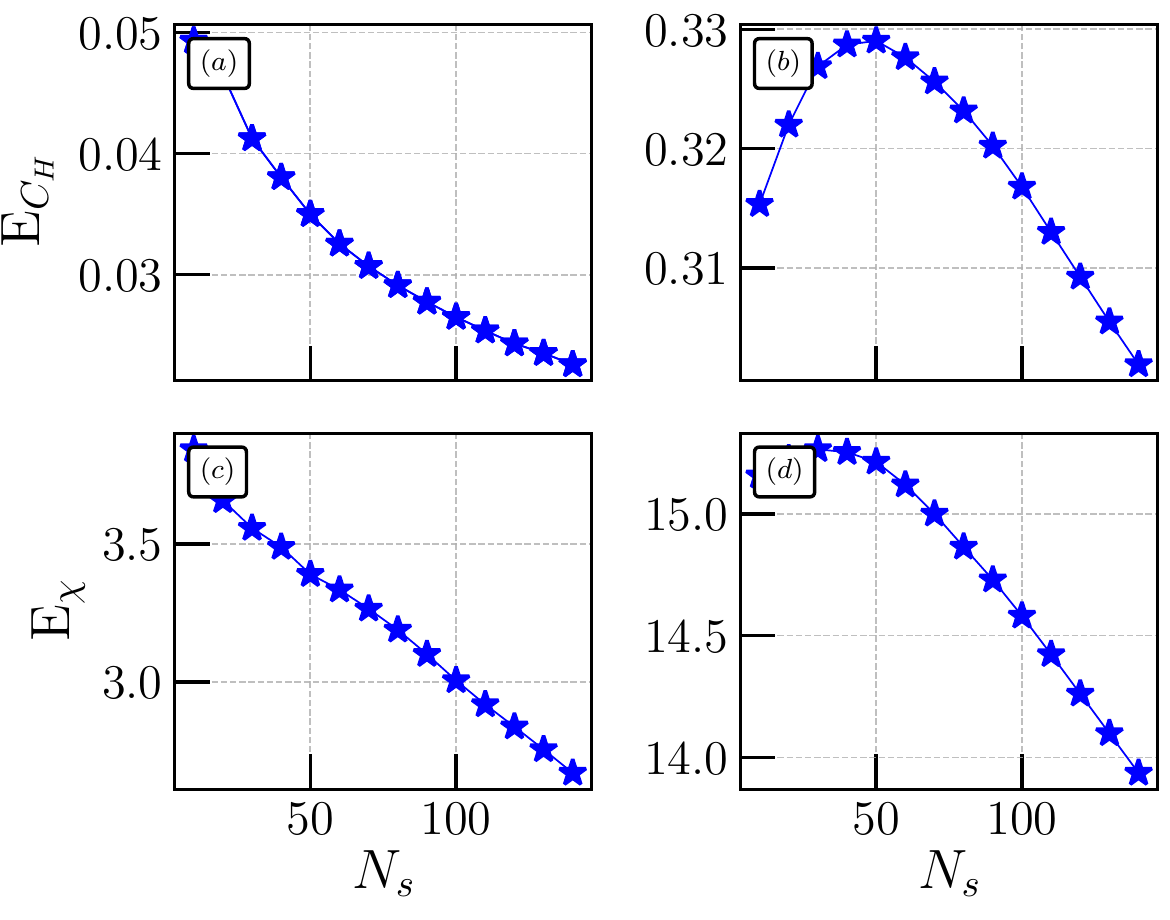}
    \caption{Mean of absolute errors (in relation to the results for MH with 1000 thermalization steps and 1000 calculation steps) for specific heat $C_H$ and susceptibility $\chi$ as functions of the number of steps $N_g$ for the $2D$ Ising model with parameters $N_I=4$, $N_P=20$ e $n_0=50$. $(a)$ and $(c)$ show the results using MPGA for thermalization with $\Delta t_{mig} =2$, $r_{mig}=0.2$ e $r_{cross}=0.6$. $(b)$ and $(d)$ show the results for MH, i.e. lacking the crossover and mutation operators on the first $n_0$ ``generations''. Temperature parameters are $\{1, 1.1,...,3.9\}$.
    }
    \label{figErrorEvoMPGAvsMC}
\end{figure}
\revfig{figEvoTopology} shows the comparison of directed graphs predicted by equations \reveq{eqCumuShapiro}, \reveq{eqCumuRede}, and \reveq{eqKl2}, corresponding to the energy minimization problem evolved by a 20-island MPGA where connections are defined by sampling of a scale free network (\revfig{figEvoTopology}$(a)$). We show that the theoretical model (\revfig{figEvoTopology}$(c)$) has good qualitative and quantitative accordance with the experimerimental result (\revfig{figEvoTopology}$(b)$). Therefore, we believe that the cumulant dynamic combined with the mapping via Kullback-Leibler divergence is an interesting tool for the study of MPGA phenomenology and for proposing better algorithms.

As previously stated, MPGA can be used as an alternative for the usual MH algorithm. To approach the 2D $20\times 20$ Ising model in absense of magnetic field, we can define a simple variant of MPGA to cover the Ising model's thermodynamics. For this purpose, we can associate each individual to a spin configuration. Where the individual fitness is given by  

$$
f^{(ln)}_\alpha = E^{(ln)}_{\alpha} = -\sum\limits_{\langle i,j\rangle}s_{i, \alpha}^{(ln)}s_{j, \alpha}^{(ln)}.
$$

The mutation operator consists in the usual mutation of the MH algorithm. Nonlinear effects in MPGA are produced by the crossover and migration operators, as explained in the appendix. Mean energy $\langle E \rangle$ and other thermodynamic quantities are recovered from the individuals' evolution in MPGA. As an example, for $n_0$ generations, the mean energy is defined as
\begin{eqnarray}
\langle E\rangle= \frac{1}{\Delta N_g}\frac{1}{N_I}\frac{1}{N_P}\sum\limits_{n=n_0 }^{N_g}\sum\limits_{l=1}^{N_I}\frac{1}{N^2}\sum_{\alpha=1}^{N_p} E_{\alpha}^{(ln)},
\label{eqMeanEMPGA}
\end{eqnarray}
and the mean magnetic moment as
\begin{eqnarray}
\langle m \rangle= \frac{1}{N_I}\frac{1}{N_P}\sum\limits_{l=1}^{N_I}\sum_{\alpha=1}^{N_p} |\frac{1}{\Delta N_g}\sum\limits_{n=n_0 }^{N_g}\frac{1}{N^2}\sum\limits_{i=1}^{L^2} s_{i, \alpha}^{(ln)}|.
\label{eqMeanMMPGA}
\end{eqnarray}
The selection of surviving individuals at each generation is given by the  following algorithm  \ref{algMetodoselecao}.
\begin{algorithm}[H]
    \SetAlgoLined
    \KwResult{Returns a new population}
    $I^{(l, n+1)} \gets \{\}$ \;
    \While{$\mathrm{size\_of}(I^{(l, n+1)}) < N_p$}{
        $\ \ \alpha \gets \text{random\_choice}( \tilde I^{(ln)}_p)$\;
        $\  \ \tilde  p^{(l,n)}_\alpha \gets \text{get\_probability\_of}(\alpha)$\;
        $\  \ \text{random} \gets \text{get\_random\_number}()$\;
        \ \ \If{$\mathrm{random} \le \tilde p^{(l,n)}_\alpha$}{
            $\  \ \  \ \text{append } \alpha \text{ into }  I^{(l, n+1)}$\;
        }
    }
    \caption{Selection Method for MPGA/MC}
    \label{algMetodoselecao}
\end{algorithm}

\revfig{figMPGAvsMC} shows a 4-island MPGA connected in a ring structure. Remaining parameters are $N_P=20$, $\Delta t_{migr} = 2$, $r_{migr}=20$, $N_g=200$. The first $50$ generations  as  used for thermalization process, The MPGA approach uses  the procedures of mutation, crossover, migration and selection. In the last $150$ generations only the MH method is applied in each individual. Using \reveq{eqMeanEMPGA} and \reveq{eqMeanMMPGA} to extract the physical quantities with $n_0=50$. Comparing results of MPGA and MH for $50$ thermalization steps and $150$ steps for calculation of the quantities, we can observe beneficial effects of evolution to ensure a better description for the MH heuristic as shown in Fig.\ref{figErrorEvoMPGAvsMC}.

\section{Conclusions} In this work we presented an extension of the theory of cumulant dynamics for MPGA. This theory combined with the proposed mapping of MPGA's islands to a graph of Kullback-Leibler divergences was shown to enable the analysis of the relation between dynamics and connectivity in MPGA. For the case of weak selection, we demonstrated that the theory describes the experimental results both qualitatively and quantitatively, elucidating the behaviour of MPGA, which can lead to the improvement of the algorithm and its parameterization. By applying MPGA to the $2D$ Ising model, we have shown that MPGA can be used as an alternative for the thermalization phase in the Metropolis–Hastings algorithm, achieving convergence in significantly fewer steps.

Note that our method applies the Gram-Charlier expansion to derive a probability distribution, which is not always possible. And although Kullback–Leibler divergence is widely used, it is not restricted by upper bound. As future work, we suggest the study of relations between topological properties given by the matrix $A$, such as reciprocity, and the dynamic of the network's properties given by the matrix $\mathbb{KL}(p_{ln}|| p_{qn})$\cite{zhang2014effects, reciprocity2017, timar2017mapping} or by the cumulants.
\begin{acknowledgments}
   \textit{Acknowledgements} -- We thank G. J. Ferreira for the suggestions and discussions provided, and also, for gently providing the computer resources employed in the experiments. We also thank T. I. de Carvalho for his advice on writing. The authors acknowledge the financial support from
   the Brazilian Agencies CNPq and CAPES.
\end{acknowledgments}

	\onecolumngrid
\appendix
\section{Mechanism for selection and migration}

For didactic reasons, we present how Shapiro obtained the equations of cumulants dynamics due to a selection process.

Let $\{f_\alpha^{(n)}\}$ be the set of fitness of a population's individuals at generation $n$. A selection process, in which the individual's ``weight''$\alpha$ is $\omega_\alpha$, has moment generating function
at generation $n+1$:

\begin{align}
M_f^{(n+1)} (t) = \sum_{\alpha=1}^{N_P} \omega_\alpha e^{t f_\alpha},
\label{eq_funcao_particao_apos_selecao}
\end{align}
It is known that the cumulants can be obtained with this function by the expression:

\begin{align}
\kappa_i^{(n+1)} = \lim_{t\to 0} \frac{\partial^i \ln M_f^{(n+1)} (t)}{\partial t^i},
\label{eq_cumulantes_particao}
\end{align}

The previous equation allows the definition of a generating function in terms of cumulants and moments ($\mu_i$),

\begin{align}
M_f^{(n+1)} (t)  =\exp\left(
    \sum\limits_{i=1}^\infty \frac{t^i\kappa_{i}^{(n+1)}}{i!}
\right) = 1 + \sum\limits_{j=1}^\infty \frac{t^i\mu_{i}^{(n+1)}}{i!} ,
\label{eqM_f^{(n+1)} (t)VsCumulantes}
\end{align}
This expression will be used later.

In the statistical analysis of GA, we associate to each individual a probability function, $p_n(f_\alpha)$ instead of fitness, which binds the probability of the individual assuming some value $f_\alpha$. Therefore, cumulants after selection must be obtained from the expected value of $\ln M_f^{(n+1)} (t)$,

\begin{align}
\kappa_i^{(n+1)} &= \lim_{t\to 0} \frac{\partial^i \langle \ln M_f^{(n+1)} (t) \rangle}{\partial t^i} \\
&= \lim_{t\to 0} \frac{\partial^i }{\partial t^i}\int\limits_{-\infty }^\infty \prod_\alpha \mathrm d f_\alpha p_n(f_\alpha) \ln M_f^{(n+1)} (t),
\label{eqCumulanteMedio1}
\end{align}
being the normalization condition

\begin{align}
\int\limits_{-\infty }^\infty\prod_\alpha \mathrm d f_\alpha p_n(f_\alpha) = 1.
\label{eqCumulanteMedioNormalizacao}
\end{align}

The integrand's logarithm at \reveq{eqCumulanteMedio1} can be represented by the following integration.

\begin{align}
\ln M_f^{(n+1)} (t) = \lim_{\epsilon\rightarrow 0}\int_\epsilon^{\infty}\mathrm d s \frac{e^{-s}- e^{-s M_f^{(n+1)} (t)}}{s}.  
\end{align}

Using the condition described at \reveq{eqCumulanteMedioNormalizacao} and applying the result above at \reveq{eqCumulanteMedio1}, we have

\begin{align}
\kappa_i^{(n+1)} &=  
    \lim_{t\to 0} \frac{\partial^i }{\partial t^i}
    \int_0^{\infty} \frac{\mathrm d s}{s}\left( e^{-s}\int\limits_{-\infty }^\infty\prod_\alpha \mathrm d f_\alpha p_n(f_\alpha)  - \int\limits_{-\infty }^\infty\prod_\alpha \mathrm d f_\alpha p_n(f_\alpha)e^{- s M_f^{(n+1)} (t)} \right)\\
&= 
    \lim_{t\to 0} \frac{\partial^i }{\partial t^i}
    \int_0^{\infty} \frac{\mathrm d s}{s}\left( e^{-s}  - \langle e^{-s M_f^{(n+1)} (t)}\rangle \right)\\
&= 
-\lim_{t\to 0} \frac{\partial^i }{\partial t^i}
\int_0^{\infty} \mathrm d s\frac{\langle e^{-s M_f^{(n+1)} (t)}\rangle}{s},
\label{eqCumulanteMedio2}
\end{align}
where $\epsilon$ and its limit is omitted to simplify the notation.

The integrand of \reveq{eqCumulanteMedio2} is given by

\begin{align}
\langle e^{-s M_f^{(n+1)} (t)}\rangle 
&=
    \int\limits_{-\infty }^\infty\prod_\alpha \mathrm d f_\alpha p_n(f_\alpha) \exp\left(-s \sum_\alpha \omega_\alpha e^{t f_\alpha}\right)
    \\
&=    
   \int\limits_{-\infty }^\infty\prod_\alpha \mathrm d f_\alpha p_n(f_\alpha) \exp\left(-s\omega_\alpha e^{t f_\alpha}\right).
\end{align}

Since there is no special order of individuals, i.e. every $p_n(f_\alpha)$ is the same, we have

\begin{align}
\langle e^{-s M_f^{(n+1)} (t)}\rangle &= \left(\int\limits_{-\infty }^\infty\mathrm d f p_n(f)\exp\left(-s\omega(f) e^{t f}\right)\right)^{N_P}\\
&=f_n(t, s)^{N_P}.
\end{align}
We can find the cumulants after a selection event by doing the following integration.

\begin{align}
\kappa_i^{(n+1)} = -\lim_{t\to 0}\frac{\partial^i}{\partial t^i}\int\limits_{0}^{\infty} \frac{\mathrm d s}{s} f_n(t,s)^{N_P}.
\label{eqCumulanteSemAproximacao}
\end{align}
Now we must get an approximation for \reveq{eqCumulanteSemAproximacao}. First we define the function

\begin{align}
\rho_n(l, t)
&=
\int\limits_{-\infty }^\infty\mathrm d f p_n(f)\left(\omega(f) e^{t f}\right)^l,
\label{eqCumulanteRho}
\end{align}

Expanding the exponential at function $f_n(t, s)$ and factoring a decreasing term with respect to $s$,

\begin{align}
f_n(t, s ) &=
    \int\limits_{-\infty }^\infty\mathrm d f p_n(f)\exp\left(-s\omega(f) e^{t f}\right) \\
    &=
        1 
        +\sum\limits_{l=1}^\infty
            \frac{(-s)^l}{l!}
            \int\limits_{-\infty }^\infty\mathrm d f p_n(f)\left(\omega(f) e^{t f}\right)^l \\
    &=
        1 
        +\sum\limits_{l=1}^\infty
            \frac{(-s)^l}{l!}
            \rho_n(l, t),
\end{align}
we get
\begin{align}
f(s, t) = 
    e^{-s  \rho_n(l, t)}
    \left[
         e^{s \rho_n(l, t)}
         \left(
            1 
            +\sum\limits_{l=1}^\infty
                \frac{(-s)^l \rho_n(l, t)}{l!}
        \right)
    \right],
\end{align}
By exponentiating the previous expression to $N_P$ and making an expansion in Taylor series,

\begin{align}
f(s, t)^{N_P} &= 
    e^{-N_P s  \rho_n(l, t)}
    \left[
        \sum\limits_{j=0}^\infty \frac{(\rho_n(l, t) s)^j}{j!}
        \left(
            1 
            +\sum\limits_{l=1}^\infty
            \frac{(-s)^l \rho_n(l, t)}{l!}
        \right)
    \right]^{N_P},
\end{align}
By neglecting terms with higher than $2$ order in $s$ inside brackets, we have

\begin{align}
f(s, t)^{N_P} &\approx 
    e^{-N_P s \rho_n(1, t)}
    \left[
        1
        \left(
            1
            +\binom{N_P}{1}(-s \rho_n(1, t) + \frac{s^2\rho_n(2, t)}{2})
            + \binom{N_P}{2} s^2\rho_n(1, t)
        \right)
    \right].
\end{align}
Terms without $\rho$ are nullified due to derivation in $t$. By substituting the previous equation in \reveq{eqCumulanteSemAproximacao}, we get

\begin{align}
\kappa_i^{(n+1)} 
    &=  
        -\lim_{t\to 0} 
            \frac{\partial^i }{\partial t^i}
                \left(
                    - \ln \rho_n(1, t) 
                    + \frac{1}{2N_P}( \rho_n(2, t) -  \rho_n(1, t)^2)
                \right).
\label{eqCumulanteP}
\end{align}
By assuming the Boltzmann selection mechanism, $\omega(f) = e^{-\beta f}$, and expanding the exponential at \reveq{eqCumulanteRho}, we get

\begin{align}
\rho(l, t) 
    &= 
        \int\limits_{-\infty }^\infty\mathrm d f p_n(f)\left(\omega(f) e^{t f}\right)^l\\
    &= 
        \sum\limits_{j=0}^\infty
            \frac{l^j(t -\beta)^j}{j!}
            \int\limits_{-\infty }^\infty\mathrm d f p_n(f) f^j\\
    &=
       1+ \sum\limits_{j=1}^\infty
            \frac{(lt -l\beta)^j}{j!}
            \mu_j^{(n)},
\end{align}
where $\mu_j^{(n)}$ is the $j$-th moment at the $n$-th generation. The previous expression represents the moment generating function (\reveq{eqM_f^{(n+1)} (t)VsCumulantes}) at point $lt-l\beta$, i.e.

\begin{align}
   \rho(l, t)  
        = 
            M_f^{(n+1)} (lt-l\beta)
        =
            \exp\left(
                \sum\limits_{j=1}^\infty \frac{(lt -l\beta)^j\kappa_{j}^{(n)}}{j!}
            \right).
\end{align}
Finally, by substituting the previous expression in \reveq{eqCumulanteP}, we get

\begin{align}
\kappa_i^{(n+1)} 
    &=  
    \lim_{u\to -\beta} 
    \frac{\partial^i }{\partial u^i}
    \left[
         \sum\limits_{j=1}^\infty \frac{u^j\kappa_{j}^{(n)}}{j!}
        - \frac{1}{2N_P}
            \exp\left(
                \sum\limits_{j=1}^\infty \frac{(2^j -2)u^j\kappa_{j}^{(n)}}{j!}
            \right)
    \right].
\end{align}
The previous expression relates cumulates at generation $n$ with cumulants at generation $n+1$ (after selection).

The required cumulants to construct the function $p_{ln}(f)$ in the event of migration must be obtained from the known formulas relating cumulants and moments. Starting from \reveq{eqCumuRede}, the second cumulant is

\begin{align}
\tilde \kappa_{2}^{(ln)}
&=\tilde \mu_2^{(ln)} - (\tilde \mu_1^{(ln)})^2\\
&=  \frac{
    n_0(\kappa_{2}^{(ln)}+(\kappa_{1}^{(ln)})^2)
    +r_{mig}\sum\limits_{j}\tilde A_{ji}(\kappa_{2}^{(jn)}+(\kappa_{1}^{(jn)})^2)+n_r\Theta(n_r)(\bar \kappa_{2}+\bar \kappa_1^2)
}
{
    n_0
    +n_m
    +n_r\Theta(n_r),
} -(\bar\kappa_{1}^{(ln)})^2,
\end{align}
And the third,
\begin{align}
\tilde \kappa_{3}^{(ln)}= \frac{
    n_0(\kappa_{3}^{(ln)}+3\kappa_{2}^{(ln)}\kappa_{1}^{(ln)}+(\kappa_{1}^{(ln)})^3)
    +r_{mig}\sum\limits_{j}\tilde A_{ji}(\kappa_{3}^{(jn)}+3\kappa_{2}^{(jn)}\kappa_{1}^{(jn)}+(\kappa_{1}^{(jn)})^3)+n_r\Theta(n_r)(\bar \kappa_{3}+\bar \kappa_1^3)
}
{
    n_0
    +n_m
    +n_r\Theta(n_r),
} \\+3(\tilde \kappa_{2}^{(ln)}+(\bar\kappa_{1}^{(ln)})^2)\bar\kappa_{1}^{(ln)}-2(\bar\kappa_{1}^{(ln)})^3.
\end{align}

\section{Kullback-Leibler divergence for Gram-Charlier expansion up to second order}

Kullback-Leibler divergence between islands $l$ and $q$ at the $n$-th generation, $\mathbb{KL}(p_{ln}|| p_{qn})
$, is defined by

\begin{align}
\mathbb{KL}(p_{ln}|| p_{qn})
	 &=  \int \mathrm d f\left[p_{ln}(f)\ln p_{ln}(f) - p_{ln}(f)\ln p_{qn}(f)\right] = - S_n(l)+ S_n(l, q),
	\label{eqKullbackLeiber}
\end{align}

where $S_n(l)$ is the entropy of the probability distribution with respect to fitness $f$ for island $l$ at the $n$-th generation; $S_n(l,q)$ is the cross-entropy between islands $l$ and $q$ in the $n$-th generation.

To simplify notation, we omit generation indexes and define that
\begin{align}
\kappa_i^{(ln)} = k_i,&
&\kappa_i^{(qn)} = q_i, &
&\Psi_{ln}(f) = \Psi(f),&
&\Psi_{qn}(f) =\Phi(f). &
\end{align}

We make a variable change $x = \frac{f-k_1}{\sqrt{k_2}} $. Also, we approximate logarithms (which contain cumulants of higher than $2$ order) up to second order terms:

\begin{align}
\ln (1+\Psi(x)) \approx \Psi(x)-\frac{\Psi(x)^2}{2} +\mathcal{O}(3).
\end{align}

Using the previous approximation, we have

\begin{align}
\label{eqEntropiaIlhaL}
S(l) = \int\mathrm d x\left[-\left(1+\Psi(x)\right)\mathcal N(x)\left(\ln \mathcal N(x)-\frac{1}{2}\ln k_2\right) 
-\mathcal N(x) \left(\Psi(x)-\frac{\Psi(x)^2}{2}\right) - \mathcal N(x)\Psi(x) \left(\Psi(x)-\frac{\Psi(x)^2}{2}\right)\right],
\end{align}

where $\mathcal N(x) = \frac{e^{-\frac{x^2}{2}}}{\sqrt 2\pi}$.

In \reveq{eqEntropiaIlhaL}, the first term's integral  is the entropy of a Gaussian distribution,

\begin{align}
-\int \mathrm d x \mathcal N(x)\left(\ln \mathcal N(x)-\frac{1}{2}\ln k_2\right)=\frac{1}{2}\ln 2\pi e k_2 .
\end{align}

The second term's integral in 
\begin{align}
 -\int \mathrm d x \mathcal N(x) \left(a_3H_3(x)+a_4H_4(x)\right)\left[-\frac{1}{2}\ln 2\pi - \frac{x^2}{2} \right],
\end{align}

where the first term inside brackets does not contribute, since $\int \mathrm d x \mathcal N(x) H_n H_0 =0$. The second term can be expanded through $x^2 = H_2(x)+H_0(x)$, as in

\begin{align}
\int \mathrm d x H_n(x)H_m(x)\mathcal N(x) = \delta_{nm}n!,
\end{align}

We can conclude that

\begin{align}
\int \mathrm d x \mathcal N(x)\Psi(x)\ln \mathcal N(x)=0 .
\end{align}

Applying the aforementioned algebric operations to the remaining terms of \reveq{eqEntropiaIlhaL} and using known expressions for integrals of Hermite polynomials, we obtain the following expression for the distribution's entropy.

\begin{align}
\label{eqEntropiaFinal}
S(l)=\frac{1}{2}\ln ( 2\pi k_2 e) -   \overbrace{\frac{k_3^2}{12 k_2^3}-\frac{k_4^2}{48 k_2^4}}^{\mathcal O(2)}+  \overbrace{\frac{k_4^3}{16 k_2^6} +\frac{3 k_3^2 k_4}{8 k_2^5}}^{\mathcal O(3)}.
\end{align}

Now we need to determine the cross term, $S(l, q)$. In this moment, it's important to emphasize that the choice for $x$ leads to $\frac{f-q_1}{\sqrt{q_2}} = \frac{x-\tilde q_1}{\sqrt{\tilde q_2}}$, where

\begin{align}
&\tilde{q_1} =  
	\frac{q_1-k_1}{\sqrt{k_2}},
	\ \ \tilde{q_2} = \frac{q_2}{k_2},
	\ \  y(x)= \frac{x-\tilde q_1}{\sqrt{\tilde q_2}},
	\ \ \mu_{q_2} = 1-\frac{1}{\tilde q_2},
	\ \ m_{Q} = -\frac{\tilde q_1}{\sqrt{\tilde q_2 -1}}.
\end{align}

The cross-entropy is then defined as

\begin{align}
S(l, q)=  -\int \mathrm{d} x\mathcal{N}(x)\left[
	1+\Psi(x)\right]\left[
		\Phi(x)-\frac{\Phi(x)^2}{2} - \frac{(x-\tilde q_1)^2}{2 \tilde q_2} -\frac{1}{2}\ln 2\pi q_2
\right] .
\label{eqcrossinicio}
\end{align}

From the orthogonality conditions of Hermite polynomials, the above equation is easily reduced to

\begin{align}
	S(l, q)=  \frac{1+\tilde q_1^2}{2\tilde q_2} + \frac{\ln 2 \pi q_2}{2}- \int \mathrm d x \mathcal N(x)\left(1+\Psi(x)\right)\left(\Phi(x)-\frac{\Phi(x^2)}{2}\right).
\end{align}

The remaining integral in above equation is not  trivial, and should be evaluated with caution.

Using known properties of Hermite polynomials and some tabled integrals \cite{gradshteyn2014table}, we can find the integrals that contribute for the cross-entropy term.

From terms which depend only of $\mathcal N(x) \Phi(x)$ appear integrals of the type

\begin{align}
\frac{1}{\sqrt{2 \pi }}\int\limits_{-\infty }^{\infty }e^{-\frac{x^2}{2}} H_n\left(y(x)\right) dx=\mu_{q_2}^{n/2} H_n(m_Q).
\end{align}

Terms with $\mathcal N(x) \Phi(x)$ contribute with integrals of type

\begin{align}
\frac{1}{\sqrt{2 \pi }}\int\limits_{-\infty }^{\infty }e^{-\frac{x^2}{2}} H_i(y(x))H_j(y(x)) dx=
\sum\limits_{k=0}^{\min (i, j)}
	k!\mu_{q_2}^{\frac{i+j-2k}{2}} \binom{j}{k}\binom{i}{k} H_{i+j-2k}(m_Q).
\end{align}

Finally, using the generating function the terms with $\mathcal N(x)\Psi(x)\Phi(x)$ contribute with integrals of type

\begin{align}
\frac{1}{\sqrt{2 \pi }}\int\limits_{-\infty }^{\infty }e^{-\frac{x^2}{2}} H_n(x)H_m(y(x)) dx=
\frac{m!}{(m-n)!}\frac{\mu_{q_2}^{\frac{m-n}{2}}}{\tilde q_2^{n/2}}H_{m-n}(-m_{\tilde Q}).
\end{align}

With knowledge of the previous integrals, the cross-entropy term is given by

\begin{align}
	\label{eqcrossentropiafinal}
	S(l, q)=  \frac{1+\tilde q_1^2}{2\tilde q_2} + \frac{\ln 2 \pi q_2}{2}
	- \overbrace{\frac{
			4 \sqrt{k_2} k_3 \left(\sqrt{q_2} q_3 \sqrt{\tilde q_2}-q_4 \tilde q_1\right)+k_4 q_4
		}{
			24 k_2^2 q_2^2 \tilde q_2^2
		}
	-\sum\limits_{j=3}^4a_j^{(q)} \mu_{q_2}^{j/2}H_j(m_Q)}^{\mathcal{O}(2)}\notag\\
	\underbrace{\frac{1}{2}\sum\limits_{i,j=3}^4
	\sum\limits_{k=0}^{\min (i, j)}
	a_i^{(l)}a_j^{(q)}k!\mu_{q_2}^{\frac{i+j-2k}{2}} 
		\binom{j}{k}\binom{i}{k} H_{i+j-2k}(m_Q)}_{\mathcal O(2)}+
	\underbrace{\frac{1}{2} \int \mathrm d x \mathcal N(x)\Psi(x)\Phi(x^2)}_{\mathcal O(3)}.
\end{align}

By neglecting terms $\mathcal O (3)$  and substituting the remaining terms in \reveq{eqKullbackLeiber}, the Kullback-Leibler divergence up to second order is defined as

\begin{align}
\mathbb{KL}(p_{ln}|| p_{qn})
&= \frac{1}{2}\ln \frac{q_2}{k_2} +\frac{k_2+ (k_1-q_1)^2}{2q_2} -\frac{1}{2}+  \mathbb{\tilde{KL}}(p_{ln}|| p_{qn})
,
\end{align}

where   $  \mathbb{\tilde{KL}}(p_{ln}|| p_{qn})$ produces the effects of third and fourth order cumulants due to Gram-Charlier expansion, which, while it can in some cases fail to represent an actual probability distribution, we believe to be sufficient to comprehend the dynamics of MPGA.

\end{document}